\documentclass[letterpaper, 10 pt, conference]{ieeeconf}

\usepackage{multicol}
\usepackage{graphicx}
\usepackage{xspace}
\usepackage{xcolor}
\usepackage{caption}

\usepackage{pifont}

\usepackage{wrapfig}
\usepackage{bbding}  
\usepackage{pifont}  
\usepackage[bookmarks=true]{hyperref}

\usepackage{booktabs}
\usepackage{float}
\usepackage{amsmath}  
\usepackage{amssymb}  
\usepackage{multirow, booktabs}
\usepackage{siunitx}

\newcommand{\ours}[0]{VisualMimic\xspace}

\newcommand{\ourwebsite}[0]{\href{https://visualmimic.github.io/
}{visualmimic.github.io}\xspace}

\definecolor{deepgreen}{RGB}{63, 126, 49}
\definecolor{deepred2}{RGB}{196, 49, 25}

\newcommand{\cmark}{\textcolor{deepgreen}{\ding{51}}}%
\newcommand{\xmark}{\textcolor{deepred2}{\ding{55}}}%

\usepackage{colortbl}
\definecolor{ourcolor}{HTML}{99e0eb}
\definecolor{ourblue}{HTML}{27a2c3}

\definecolor{tablecolor}{HTML}{ccf2f5} 

\definecolor{tablecolor2}{HTML}{ffcdb4}
\definecolor{citecolor}{HTML}{fe7b5b}
\definecolor{grey}{rgb}{0.9, 0.9, 0.9}
\usepackage{amssymb}

\usepackage{listings}
\definecolor{gred}{rgb}{0.859,0.267,0.216}
\definecolor{ggreen}{rgb}{0.059,0.616,0.345}

\definecolor{deepblue}{HTML}{27a2c3}

\definecolor{deepred}{HTML}{7c2320}

\usepackage{cite}
\usepackage{amsmath,amssymb,amsfonts}
\usepackage{algorithmic}
\usepackage{graphicx}
\usepackage{textcomp}
\usepackage{xcolor}
\usepackage{subcaption}
\begin{document}
\definecolor{myblue}{RGB}{215,233,242}
\newcommand{\karen}[1]{\textbf{\color{red}\small [Karen: #1]}}
\newcommand{\shaofeng}[1]{\textbf{\color{blue}\small [Shaofeng: #1]}}
\newcommand{\jw}[1]{\textbf{\color{green}\small [Jiajun: #1]}}

\title{

\textbf{\ours}\\ 
Visual Humanoid Loco-Manipulation via Motion Tracking and Generation

}

\author{
Shaofeng Yin*\quad Yanjie Ze*\quad Hong-Xing Yu\quad C. Karen Liu$^{\dag{}}$\quad Jiajun Wu$^{\dag{}}$\\
 *Equal Contribution\quad $^{\dag{}}$Equal Advising\\
\textbf{Stanford University}\\
}

\twocolumn[{%
\renewcommand\twocolumn[1][]{#1}%
\maketitle
\vspace{-0.2in}
\begin{center}
    \centering
    \includegraphics[width=1.0\linewidth]{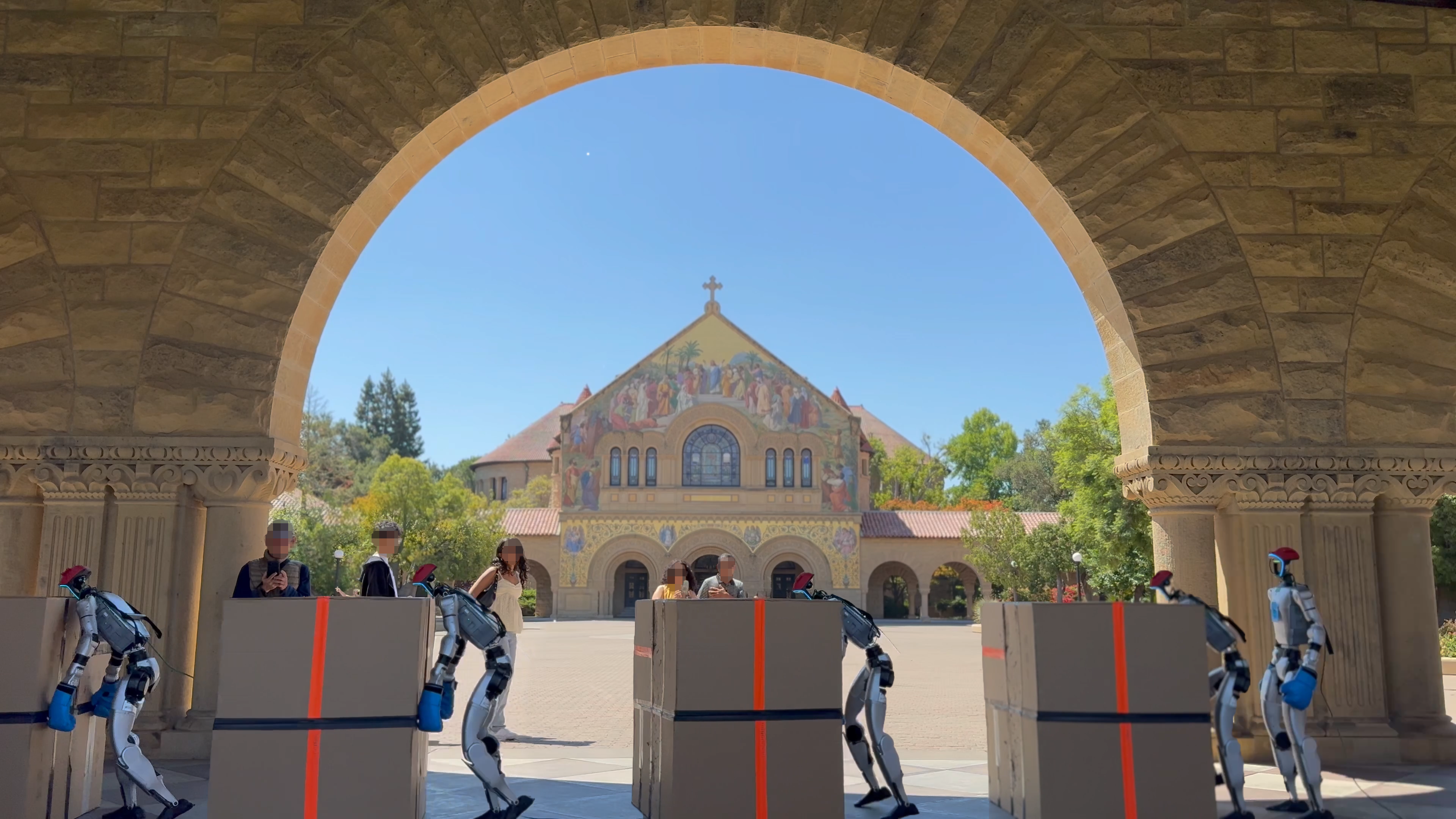}
    \captionsetup{type=figure}
    \caption{We present \ours, a visual sim-to-real framework for whole-body humanoid loco-manipulation. 
    Videos are available at \ourwebsite.
    }
    \label{fig:push_box}
\end{center}
}]

\begin{abstract}
\footnotetext[1]{Work was done during Shaofeng Yin's internship at Stanford University. Shaofeng is now with Tsinghua University.}
Humanoid loco-manipulation in unstructured environments demands tight integration of egocentric perception and whole-body control. However, existing approaches either depend on external motion capture systems or fail to generalize across diverse tasks. We introduce VisualMimic, a visual sim-to-real framework that unifies egocentric vision with hierarchical whole-body control for humanoid robots. VisualMimic combines a task-agnostic low-level keypoint tracker---trained from human motion data via a teacher-student scheme---with a task-specific high-level policy that generates keypoint commands from visual and proprioceptive input. To ensure stable training, we inject noise into the low-level policy and clip high-level actions using human motion statistics. VisualMimic enables zero-shot transfer of visuomotor policies trained in simulation to real humanoid robots, accomplishing a wide range of loco-manipulation tasks such as box lifting, pushing, football dribbling, and kicking. Beyond controlled laboratory settings, our policies also generalize robustly to outdoor environments. 
Videos are available at: \ourwebsite

\end{abstract}

\section{Introduction}

How do humans manage to push a box that is too heavy to move with only their arms? We start with vision perception to localize the box and rely on visual feedback to guide our interaction with the box. To generate sufficient force, we might bend down and push with our hands, lean in with the strength of our arms and shoulders, or even nudge the box forward with our feet. In such cases, every part of the body can be brought into play to accomplish the task. These strategies underscore two fundamental aspects of human loco-manipulation: egocentric visual perception and whole-body dexterity.

Equipping humanoid robots with such human-like object interaction abilities has been a long-standing challenge. Current approaches can be categorized into three main paradigms based on tasks: First, locomotion-focused methods~\cite{RealHumanoid2023,TokenHumanoid2024} that excel at terrain traversal but do not address object interaction. Second, approaches that rely on external motion capture systems~\cite{su2025hitter,xue2025full} for object state estimation, restricting their deployment to controlled laboratory environments. Third, vision-based methods for object interaction, which follow two distinct paths: 1) imitation learning approaches~\cite{ze2025twist,ben2024homie,li2025amo} that train visuomotor policies via human demonstrations, which are constrained by the scarcity of large-scale demonstration data and result in limited generalization capabilities; and 2) sim-to-real reinforcement learning (RL) methods~\cite{videomimic,wang2025beamdojo} that offer greater robustness and generalizability; however, vision-based RL is currently limited to simple environmental interactions such as sitting~\cite{videomimic} and stair climbing~\cite{videomimic,wang2025beamdojo}, falling significantly short of human-level object interaction abilities, due to the large exploration and action space of humanoid robots.


We aim to take one step forward on the pathway of sim-to-real RL for visual humanoid-object interaction. To make sim-to-real RL generalize better, we adopt a hierarchical design comprising low-level and high-level policies. In such a hierarchical framework, the task-agnostic low-level policy takes care of balanced control and tracks the command sent by the high-level policy, and the task-specific high-level policy generates simplified tracking commands conditioning on egocentric vision input. This design enables more effective task-specific training. We formulate the command interface as body keypoints (root, hands, feet, head) to ensure both compactness and expressiveness. 

To obtain a low-level keypoint tracker that performs human-like behaviors while tracking commands, we curate human motion data and supervise the tracker via motion imitation rewards. However, because keypoint commands alone do not capture the entirety of human motion,  we observe that the keypoint tracker can track target keypoints while not perfectly producing human-like behaviors. To address this problem, we adopt a teacher–student training scheme: 1) We first train a motion tracker with full access to current and future whole-body motions, thereby capable of precisely following human reference motions; 2) We then distill this motion tracker into a keypoint tracker that operates on simplified keypoint commands. By doing so, our keypoint tracker captures human motion behaviors while still maintaining a compact command space. Notably, our keypoint tracker is task-agnostic and shared across tasks once trained.

Built upon this general keypoint tracker, we train a high-level keypoint generator via sim-to-real RL. Directly training polices via visual RL significantly slows down the training and leads to non-optimal solutions. Therefore, we also apply a teacher–student scheme: 1) We first train a state-based policy with privileged access to object states, enabling them to solve tasks effectively; 2) We then distill the state-based policy into the visuomotor policy that rely solely on egocentric vision and robot proprioception, making it ready for real-world deployment without external object state estimation. To address the large visual sim-to-real gap (Fig. \ref{fig:vision}), we apply heavy masking to depth images in simulation, approximating real-world sensor noise. 

Due to the exploration nature of RL, we find that the high-level policy training is not stable when the high-level policies explore the action space that is beyond the human motion space (HMS) present in training motion datasets. We adopt two strategies to alleviate this problem: 1) injecting noise during training the low-level policy to help it adapt to potentially noisy commands from the high-level policy, and 2) clipping actions from the high-level policy to keep them within the feasible HMS.

The resulting framework, \ours, enables us to obtain robust and generalizable visuomotor policies that can zero-shot transfer to the real robot, across a broad range of humanoid loco-manipulation tasks, with relatively simple task-specific reward design and without requiring paired human-object motion data. For real-world experiments (Fig. \ref{fig:real_world} and Fig. \ref{fig:push_box_real}), we show that our humanoid robot can 1) lift a 0.5-kilogram box to a height of 1 meter, 2) push a very large box (similar height as the robot and weight 3.8 kilograms) straight and steady with its whole body, 3) dribble a football with the fluency of an experienced player, and 4) kick a box forward with alternating feet.  Notably, we also show that our visuomotor policies achieve stable performance in outdoor scenarios, showing strong robustness to real-world variability such as lighting changes and uneven ground.
\section{Related Work}

\begin{figure*}[t]
    \centering
    \includegraphics[width=0.9\linewidth]{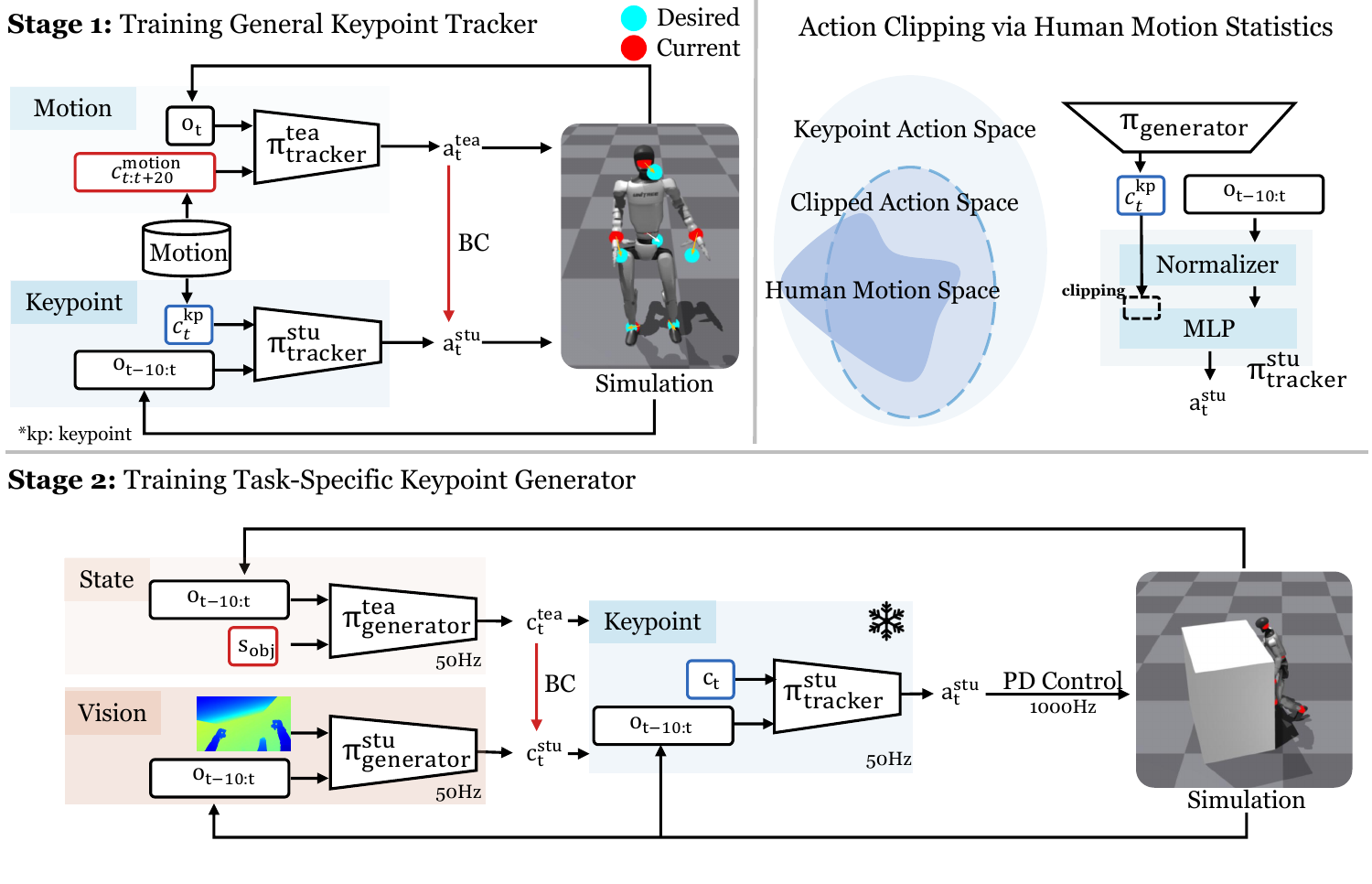}
    \captionsetup{type=figure}
    \vspace{-5pt}
    \caption{\ours consists of two training stages: 1) \textbf{training a general keypoint tracker}, where a teacher motion tracker is first trained and then distilled into a keypoint tracker with keypoint commands; and (2) \textbf{training a task-specific keypoint generator}, where a teacher policy with privileged object states is first trained and then distilled into a visuomotor policy. To ensure stable learning, we compute statistics with human motions and use them to clip high-level actions. Here, $o_t$ is the proprioceptive observation at time $t$, $a_t$ is the action, and $s_{\text{obj}}$ represents the object state.} 
    \label{fig:method}
    \vspace{-10pt}
\end{figure*}

\begin{table}[tbp]
  \centering
  \caption{Comparison of methods of different features.}
  \resizebox{1.0\linewidth}{!}{%
    \begin{tabular}{lccc}
      \toprule
      \textbf{Method} & \textbf{Whole-Body Dex} & \textbf{Loco-Manipulation} &\textbf{Visual Policy} \\
      \midrule
      TWIST \cite{ze2025twist}  & \cmark & \cmark &\xmark \\
      VideoMimic \cite{videomimic}  & \xmark & \xmark  & \cmark \\
      Hitter \cite{su2025hitter}  & \xmark & \cmark  & \xmark \\
      Recipe \cite{lin2025sim}  & \xmark & \xmark  & \cmark \\

      HEAD \cite{head}  & \xmark & \xmark  & \cmark \\
      \midrule
      \textbf{\ours (Ours)} & \cmark & \cmark & \cmark\\
      \bottomrule
    \end{tabular}
  }
  \label{tab:comparison}
\end{table}

\subsection{Learning Humanoid Loco-Manipulation}

Enabling humanoid robots to perform versatile loco-manipulation in unstructured environments, akin to humans, is a long-standing goal for roboticists. Currently, two main pathways are being explored: (1) Imitation learning on real-world data collected via whole-body teleoperation~\cite{ben2024homie,li2025amo,ze2025twist,he2024omnih2o,fu2024humanplus}. While these methods demonstrate promising task versatility, they remain limited by the scarcity of high-quality data and the difficulty of scaling data collection. (2) Sim-to-real reinforcement learning based on large-scale simulation interaction~\cite{videomimic,su2025hitter,sferrazza2024humanoidbench,chen2024lcp,wang2025beamdojo,dao2024sim}. These approaches exhibit strong generalization in specific humanoid motor skills, such as terrain traversal~\cite{wang2025beamdojo,chen2024lcp}, box picking~\cite{dao2024sim}, and table tennis~\cite{su2025hitter}, but are restricted in task diversity compared to imitation learning. Several works remain confined to simulation; for example, HumanoidBench~\cite{sferrazza2024humanoidbench} and SkillBlender~\cite{kuang2025skillblender} adopt hierarchical frameworks similar to ours. However, their policies are often excessively jittery or depend on privileged object states, hindering successful deployment in the real world.

Previous works on learning visuomotor policies for humanoid robots have typically focused either on upper-body manipulation~\cite{ze2024idp3,cheng2024tv,qiu2025humanoid,lin2025sim} or on perceptive locomotion~\cite{wang2025beamdojo,He2025AttentionBased,long2024learning}. More recently, VideoMimic~\cite{videomimic} introduced a real2sim2real pipeline that enables real robots to perform environment interactions such as sitting, though their interactions remain limited to static settings like the ground or stone chairs. Other efforts, such as PDC~\cite{luo2025emergent}, have demonstrated promising results in vision-based whole-body loco-manipulation, but only in simulation. In contrast, we propose a sim-to-real framework that enables real-world humanoid robots to achieve versatile object interaction and loco-manipulation through egocentric vision.

\begin{figure*}[tpb]
    \centering
    \includegraphics[width=1.0\linewidth]{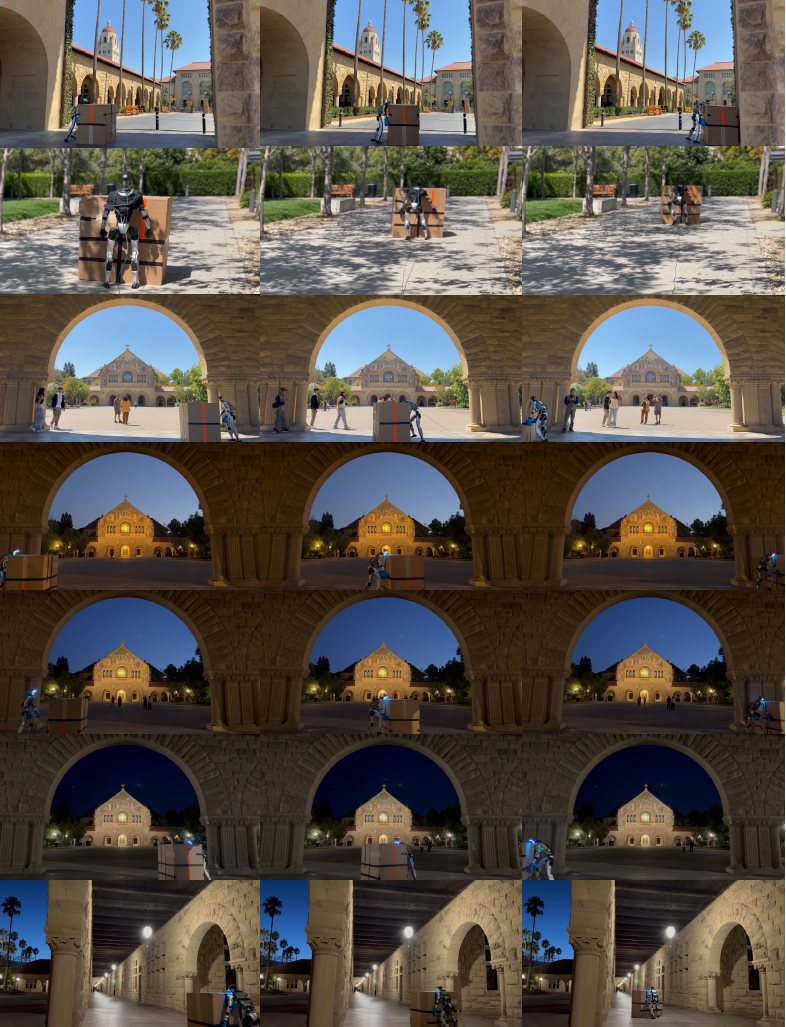}
    \captionsetup{type=figure}
    \caption{Our visuomotor policies generalize across diverse space and time, shown on the box-pushing task.} 
    \label{fig:push_box_real}
\end{figure*}

\section{Method}

\begin{figure*}[t]
    \centering
    \includegraphics[width=1.0\linewidth]{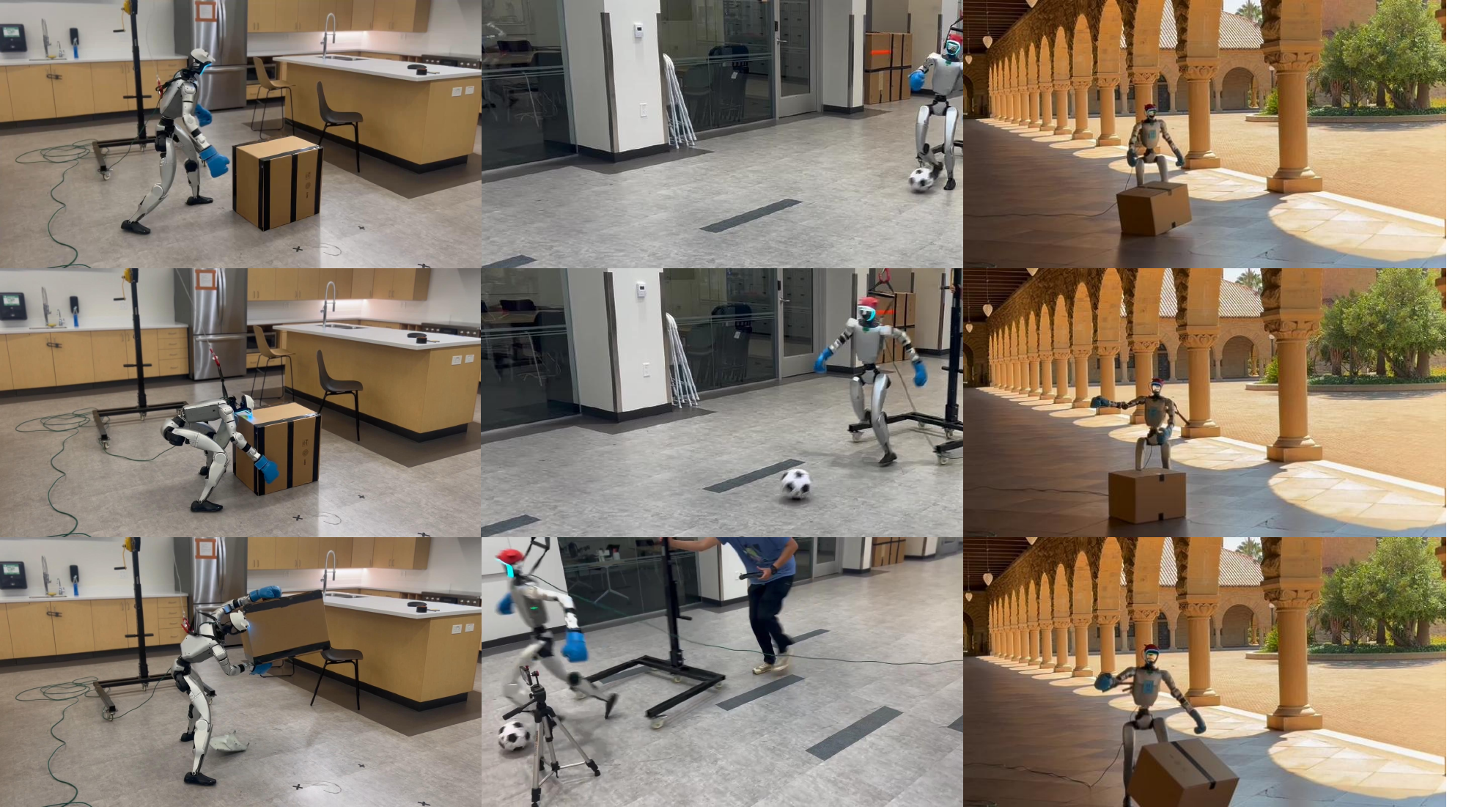}
    \captionsetup{type=figure}
    \caption{Real-world deployment of visuomotor policies on a humanoid, showcasing diverse loco-manipulation tasks: Lift Box, Kick Ball, and Kick Box. }
    \label{fig:real_world}
\end{figure*}

We propose \ours, a sim-to-real framework for real-world humanoid loco-manipulation by integrating both egocentric visual perception and whole-body dexterity priors within a hierarchical framework (see Fig.~\ref{fig:method} for overview). Our approach consists of two main components: 1) a low-level task-agnostic keypoint tracking policy $\pi_{\text{tracker}}$ that learns whole-body dexterity priors from human motion data, and 2) a high-level task-specific visuomotor policy $\pi_{\text{generator}}$  that drives the low-level policy based on egocentric visual inputs (Section~\ref{sec:keypoint generator}). Both policies are trained purely via large-scale simulation and zero-shot transfer to the real robot. This hierarchical design facilitates fast adaptation to new tasks, since only the high-level policy requires training per task. We detail the design of two policies in this section.

\subsection{General Keypoint Tracker}
\label{sec:general_keypoint_tracker}

Although a keypoint tracking policy can be trained directly, its ability to capture motion is weaker than that of a motion tracking policy due to the much simplified commands, resulting in less human-like behaviors (Fig. \ref{fig:kick_vis}). We address this problem by designing a two-stage teacher-student distillation pipeline.  Specifically,  in the teacher training stage, a privileged teacher motion tracker is trained via RL with access to future reference motions. Then, a student keypoint tracker is trained using DAgger \cite{ross2011reduction}, relying solely on proprioception and keypoint commands computed from the reference frame at each time step.

\paragraph{Teacher Motion Tracker} 

Since the teacher policy is not used during deployment, we provide it with sufficient motion and proprioceptive information so that it can track as accurately as possible. The teacher motion tracker $\pi_\text{tracker}^{\text{tea}}$  takes as input a sequence of future reference motion frames (spanning 2 seconds) and privileged proprioceptive information (e.g., foot contact forces), which allows it to anticipate upcoming goals and generate smoother motion. We implement $\pi_\text{tracker}^{\text{tea}}$ as a simple three-layer MLP and optimize it using PPO~\cite{schulman2017ppo,rudin2022legged_gym}. Following the reward structure in \cite{ze2025twist}, our reward $r_\text{motion}$ encourages accurate motion tracking while penalizing artifacts such as jitter and foot slippage: $r_\text{motion} = r_\text{track} + r_\text{penalty}$. Notably, we track robot body positions and root velocities in the world frame. For motion datasets, we use GMR~\cite{ze2025gmr,ze2025twist} to retarget AMASS~\cite{mahmood2019amass} and OMOMO~\cite{li2023omomo} into humanoid motions.

\paragraph{Student Keypoint Tracker}  

After obtaining $\pi_\text{generator}^{\text{tea}}$, we distill it into $\pi_\text{generator}^{\text{stu}}$ via DAgger~\cite{ross2011reduction}, which takes keypoint commands $c^{\text{kp}}_t$ as input and is real-world deployable. We define
\begin{equation}
c^{\text{kp}}_t = \bigl[\Delta p_t,\, \Delta x_t^1,\, \ldots,\, \Delta x_t^5 \bigr],
\end{equation}
where the root position error is
\begin{equation}
\Delta p_t = p^{\text{des}}_t - p^{\text{cur}}_t,
\end{equation}
and the keypoint errors (for head, two hands, and two feet) are
\begin{equation}
\Delta x_t^i = \bigl(x^{i,\text{des}}_t - p^{\text{des}}_t\bigr) - \bigl(x^{i,\text{cur}}_t - p^{\text{cur}}_t\bigr), \quad i=1,\dots,5.
\end{equation}
Here, “des” and “cur” denote desired (reference) and current trajectories. The keypoint tracker $\pi_\text{tracker}^{\text{stu}}$ relies only on proprioception and the immediate command $c^{\text{kp}}_t$. Following the teacher tracker, it is implemented as a three-layer MLP.

\subsection{Task-Specific Keypoint Generator}
\label{sec:keypoint generator}

Once the low-level keypoint tracker is trained, the next step is to develop a high-level keypoint generator that directs the tracker to perform diverse tasks. 
Benefiting from the low-level tracker trained with human motion, we only need to focus on guiding the robot toward task completion by designing a small set of task rewards, without the additional burden of ensuring human-like motion or collecting pairwise human-object interaction data. 

Directly training such a keypoint generator with visual RL, however, is highly inefficient (Table~\ref{tab:ablation}), because crucial information—such as object positions and contact forces—is only partially observable, and incorporating vision into IsaacGym further slows down the simulation. To overcome these challenges, we adopt a two-stage approach to train a task-specific keypoint generator. In the first stage, a teacher generator with access to task-relevant object states is trained using PPO \cite{schulman2017ppo}. In the second stage, a student keypoint tracker that relies only on depth images for object information is distilled.

\paragraph{Teacher State-based Keypoint Generator}

As discussed in the previous paragraph, the teacher keypoint generator leverages object states to accelerate training. The object state is defined within the environment and concatenated with proprioceptive information to form the input of the state-based keypoint generator. The generator, implemented as a three-layer MLP, is trained using PPO~\cite{schulman2017ppo,rudin2022legged_gym} with a task-specific reward function.

We focus on loco-manipulation tasks such as pushing/reaching/kicking objects. Our task rewards are as follows:
\begin{enumerate}
\item \textbf{Approach} ($R_{\mathrm{approach}}$): encourages contact with target point(s) on target objects.  
For single-point contact:
\begin{equation}
R_{\mathrm{approach}}(t)=e^{-0.1 d(t)},
\end{equation}
and for two-point contact such as pushing a box with two hands, we use a harmonic mean for balance:
\begin{equation}
R_{\mathrm{approach}}(t)
= \frac{2\,e^{-0.1 d_1(t)}e^{-0.1 d_2(t)}}{e^{-0.1 d_1(t)}+e^{-0.1 d_2(t)}},
\end{equation}
where $d(t)$ (or $d_1(t), d_2(t)$) denotes the distance between a humanoid end-effector (hand or foot) and the target point(s) at time $t$.
\item \textbf{Forward progress} ($R_{\mathrm{forward}}$): rewards new forward motion of the object:
\begin{equation}
R_{\mathrm{forward}}(t)
= \tanh\!\Bigl(10[x_{\text{obj}}(t)-\max_{t'<t}x_{\text{obj}}(t')]_+\Bigr),
\end{equation}
where $x_{\text{obj}}(t)$ denotes the object's position coordinate along the forward direction.
\item \textbf{Force} ($R_{\mathrm{force}}$): rewarding sufficient force exerted on the object:
\begin{equation}
R_{\mathrm{force}}(t)
= e^{-0.1[F_{\text{des}}-F_{\text{obj}}(t)]_+},
\end{equation}
where
$F_{\text{obj}}(t)$ is the force exerted on the object and $F_{\text{des}}$ is the desired force threshold.
\end{enumerate}

Besides, we have the following terms on regularizing the task behavior of the policy: 
\begin{enumerate}
\item \textbf{Look at object} ($R_{\mathrm{look}}$): encourages the robot to face the object:
\begin{equation}
R_{\mathrm{look}}(t)
= -\bigl(\arccos(\hat{\mathbf{f}}_{\text{body}}\cdot\hat{\mathbf{d}}_{\text{obj}})\bigr)^2 .
\end{equation}
$\hat{\mathbf{f}}_{\text{body}}$ denotes the unit vector of the humanoid's facing direction, and 
$\hat{\mathbf{d}}_{\text{obj}}$ is the unit vector pointing from the humanoid toward the target object.
\item \textbf{Drift} ($R_{\mathrm{drift}}$): penalizes lateral deviation:
\begin{equation}
R_{\mathrm{drift}}(t)
= -\,\tanh\!\Bigl(10[\,|y_{\text{obj}}(t)|-\max_{t'<t}|y_{\text{obj}}(t')|\,]_+\Bigr).
\end{equation}
$y_{\text{obj}}(t)$ denote the object's position components along the lateral direction.
\end{enumerate}

\paragraph{Student Vision-based Keypoint Generator}

Since object states are unavailable during deployment, the state-based keypoint generator $\pi_\text{generator}^{\text{tea}}$ is distilled into a student keypoint generator that relies solely on visual observations and proprioceptive inputs. Since RGB images have significant sim-to-real gaps, we use depth images as the only visual modality. The depth input is processed by a CNN encoder, whose output is then concatenated with proprioceptive features and passed through an MLP. The student keypoint generator is distilled using DAgger \cite{ross2011reduction}.

\begin{figure*}[tbp]
    \centering
    \includegraphics[width=1.0\linewidth]{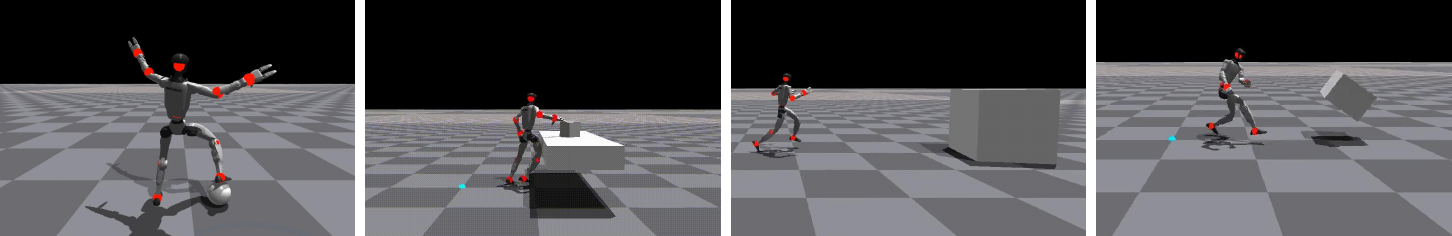}
    \captionsetup{type=figure}
    \caption{Visuomotor policies perform diverse loco-manipulation tasks in simulation: from left to right, Balance Ball, Push Cube, Reach Box, Large Kick.} 
    \label{fig:sim}
\end{figure*}

\subsection{Clipping Action Space to Human Motion Space}

We find that it is hard to maintain the training stability of $\pi_{\text{generator}}$ even though we use the compact command space, because RL needs heavy exploration during training and the exploration easily goes beyond a feasible space of keypoint commands extracted from human motions. We refer to this feasible space as the \textbf{Human Motion Space} (HMS). To alleviate the problem of action exploration beyond the HMS, we propose the following two techniques.

\paragraph{Noised Keypoint Commands for Low-Level Student Training}
\label{method:noise}
To enhance the robustness of the low-level policy and expand the range of Human Motion Space, we inject multiplicative noise into each dimension of the keypoint command during training. Formally, the noised command is defined as
$X_{\text{noised}} = X \cdot \lambda_i, 
\quad \forall i \in \{1, \dots, n\}, 
\quad \lambda_i \sim \mathcal{U}(0.5, 1.5)$, where $X$ denotes the original command and $\lambda_i$ is sampled independently from a uniform distribution. We set the relative noise level to 50\%, which is large enough to diversify keypoint commands while preserving motion signals. Empirically, this strategy significantly benefits the subsequent training of the keypoint generator (Fig. \ref{fig:ablate_noise}). 

\paragraph{Action Clip for High-Level Policies}
\label{method:clip}
Besides injecting noise for increasing robustness, we further regularize the output of $\pi_{\text{generator}}$. To this end, we first estimate the the HMS boundary using the low-level policy input normalizer, and then apply action clip to constrain the high-level policy output within this range. Specifically, each input dimension is modeled as a Gaussian distribution, and the feasible output range for high-level policy is defined as $\mu \pm 1.64\sigma$, which covers approximately 90\% of the probability mass. The mean $\mu$ and standard deviation $\sigma$ are recorded during low-level policy training. Fig. \ref{fig:ablate_clip} shows that action clip significantly stabilizes the training of the keypoint generator.

\subsection{Real World Deployment}

\label{ref:real_world_deployment}

\paragraph{Vision-Based Sim-to-Real Transfer}

\begin{figure}[tbp]
    \centering
    \includegraphics[width=1.0\linewidth]{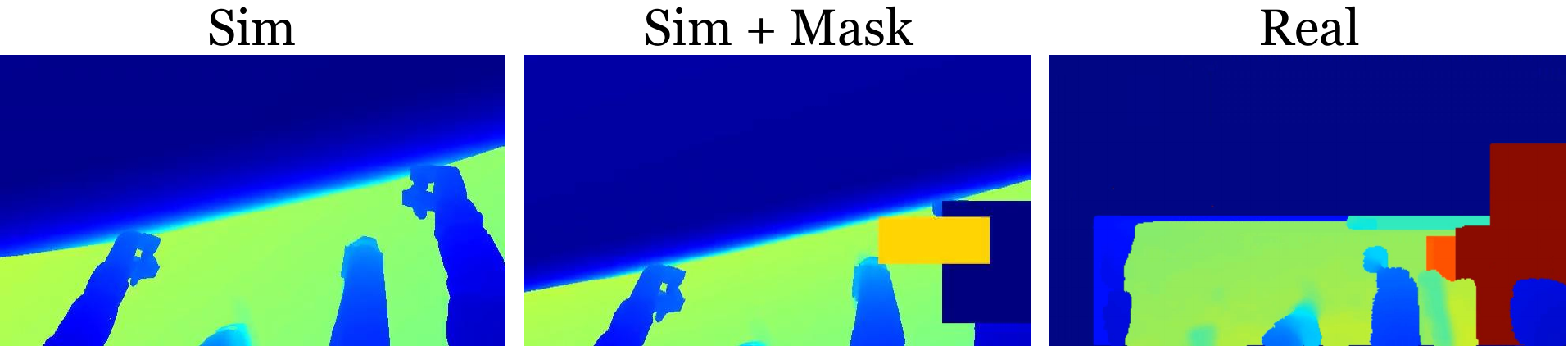}
    \captionsetup{type=figure}
    \vspace{-10pt}
    \caption{Egocentric vision of the humanoid robot. In simulation, random masking is applied to approximate real-world noise. Real-world images are processed with spatial and temporal filtering.}
    \label{fig:vision}
    \vspace{-10pt}
\end{figure}

We observe that the depth images from the RealSense camera are highly noisy. To mitigate this, we apply spatial and temporal filters to smooth the real-world depth images. As shown in Fig.~\ref{fig:vision}, even after smoothing, a significant gap remains between simulated and real-world depth images. To address this issue, we apply heavy random masking during training to better approximate real-world visual noise. Specifically, with 20\% probability we apply a fixed bottom-left white mask, and with 10\% probability each we add up to six independently sampled rectangular masks. These masks are filled with white, black, or gray values, where gray is drawn uniformly from 0 to 1. Each mask covers up to 30$\times$30 pixels on an 80$\times$45 frame (25\% of the image).  Without such masking, the robot exhibits unstable behaviors during deployment. We further notice that the angle of RealSense camera on Unitree G1 has slight drift as the neck is not fixed stably. To account for this effect, we apply randomization on the orientation of the robot camera view by up to $\pm 5^\circ$.

\paragraph{Safe Real-World Deployment via Binary Commands}

In real-world deployment, it is crucial for the robot to start/pause/end safely during task execution, as simply terminating the program may cause it to fall and get damaged. Therefore, we introduce a binary command signal (0 or 1) that instructs the robot to either pause or execute the task. The robot can freely switch between the two states and always starts in the pause state. We train this behavior with such reward design: when the command is 0, the task reward is disabled, and when the command is 1, the pause reward is disabled. The pause reward corresponds to tracking a stationary standing motion. Both commands are sampled with 50\% probability.
\section{Experiments and Analysis}

\begin{table*}[tbp]
\centering
\caption{Evaluation of teacher and student policies with and without vision across all tasks in simulation. Results are averaged over 1-minute rollouts across 3 seeds (4096 rollouts for state-based, 512 for vision-based).
Metrics include: \textbf{Distance} (object Euclidean displacement), \textbf{Forward} (object displacement in the intended forward direction), \textbf{Drift} (object lateral deviation orthogonal to the forward axis), \textbf{Height} (box lift height), \textbf{Box Fall Rate} (percentage of drops),   \textbf{Alive} (time before termination), \textbf{Velocity} (forward speed),  
\textbf{Collision Rate} (percentage of collisions), \textbf{Force} (average force applied on the ball), 
\textbf{Foot Fall Rate} (percentage of failed foot placements),  \textbf{Error} (final distance from target), and \textbf{Finish Time} (time to task completion; set to episode length if the task is not completed).
\textbf{The best performing deployable method is highlighted.}
} 
\resizebox{\textwidth}{!}{
\begin{tabular}{l|ccc|ccc|ccc|ccc}
\toprule
\textbf{Method} 
& \multicolumn{3}{c|}{\textbf{Push Box}} 
& \multicolumn{3}{c|}{\textbf{Kick Box}} 
& \multicolumn{3}{c|}{\textbf{Lift Box}} 
& \multicolumn{3}{c}{\textbf{Reach Box}} \\
\cmidrule(lr){2-4}\cmidrule(lr){5-7}\cmidrule(lr){8-10}\cmidrule(lr){11-13}
& Distance [m] $\uparrow$ & Forward [m] $\uparrow$ & Drift [m] $\downarrow$
& Distance [m] $\uparrow$ & Forward [m] $\uparrow$ & Drift [m] $\downarrow$
& Height [m] $\uparrow$ & Box Fall Rate [\%] $\downarrow$ & Alive [s] $\uparrow$
& Velocity [m/s] $\uparrow$ & Collision Rate [\%] $\downarrow$ & Alive [s] $\uparrow$ \\
\midrule
teacher & 152 ± 36 & 151 ± 29 & 13 ± 4 & 78 ± 3 & 78 ± 3 & 0 ± 0 & 1 ± 0 & 34 ± 25 & 38 ± 13 & 4 ± 0 & 0 ± 0 & 60 ± 0 \\
stu w/ vision & \cellcolor{myblue}\textbf{37 ± 28} & \cellcolor{myblue}\textbf{19 ± 15} & 21 ± 12 & \cellcolor{myblue}\textbf{55 ± 5} & \cellcolor{myblue}\textbf{30 ± 3} & 33 ± 3 & \cellcolor{myblue}\textbf{1 ± 0} & 30 ± 23 & \cellcolor{myblue}\textbf{30 ± 7} & \cellcolor{myblue}\textbf{4 ± 0} & \cellcolor{myblue}\textbf{0 ± 0} & \cellcolor{myblue}\textbf{42 ± 6} \\
stu w/o vision & 2 ± 0 & 2 ± 0 & \cellcolor{myblue}\textbf{1 ± 0} & 0 ± 0 & 0 ± 0 & \cellcolor{myblue}\textbf{0 ± 0} & 0 ± 0 & \cellcolor{myblue}\textbf{15 ± 21} & 6 ± 4 & \cellcolor{myblue}\textbf{4 ± 0} & \cellcolor{myblue}\textbf{0 ± 0} & 18 ± 6 \\
\midrule
\textbf{Method} 
& \multicolumn{3}{c|}{\textbf{Large Kick}} 
& \multicolumn{3}{c|}{\textbf{Kick Ball}} 
& \multicolumn{3}{c|}{\textbf{Balance Ball}} 
& \multicolumn{3}{c}{\textbf{Push Cube (Tabletop)}} \\
\cmidrule(lr){2-4}\cmidrule(lr){5-7}\cmidrule(lr){8-10}\cmidrule(lr){11-13}
& Distance [m] $\uparrow$ & Forward [m] $\uparrow$ & Drift [m] $\downarrow$
& Distance [m] $\uparrow$ & Forward [m] $\uparrow$ & Drift [m] $\downarrow$
& Force [N] $\uparrow$ & Foot Fall Rate [\%] $\downarrow$ & Alive [s] $\uparrow$
& Error [cm] $\downarrow$ & Finish Time [s] $\downarrow$ & Alive [s] $\uparrow$ \\
\midrule
teacher & 8 ± 1 & 7 ± 1 & 2 ± 0 & 189 ± 3 & 189 ± 3 & 4 ± 1 & 21 ± 2 & 0 ± 0 & 34 ± 8 & 9 ± 3 & 4 ± 1 & 58 ± 1 \\
stu w/ vision & \cellcolor{myblue}\textbf{6 ± 0} & \cellcolor{myblue}\textbf{6 ± 0} & 2 ± 0 & \cellcolor{myblue}\textbf{135 ± 6} & \cellcolor{myblue}\textbf{121 ± 8} & 47 ± 12 & \cellcolor{myblue}\textbf{24 ± 1} & \cellcolor{myblue}\textbf{0 ± 0} & \cellcolor{myblue}\textbf{45 ± 7} & \cellcolor{myblue}\textbf{21 ± 2} & \cellcolor{myblue}\textbf{20 ± 8} & \cellcolor{myblue}\textbf{57 ± 0} \\
stu w/o vision & 4 ± 0 & 4 ± 0 & \cellcolor{myblue}\textbf{1 ± 0} & 1 ± 0 & 1 ± 0 & \cellcolor{myblue}\textbf{0 ± 0} & 6 ± 0 & \cellcolor{myblue}\textbf{0 ± 0} & 5 ± 1 & 57 ± 22 & 43 ± 8 & 51 ± 10 \\
\bottomrule
\end{tabular}
}
\label{tab:sim}
\end{table*}

In this section, we perform a series of experiments aimed at addressing the following questions:
\begin{itemize}
    \item[] Q1: Does \ours enable effective training of humanoids to perform diverse tasks in a human-like manner?
    \item[] Q2: Can policies trained with \ours transfer robustly to the real world?  
    \item[] Q3: How effectively does our framework demonstrate whole-body dexterity?
    \item[] Q4: How well does \ours utilize vision for object interaction?
    \item[] Q5: Are design choices of \ours all necessary for the success of the system?
\end{itemize}

\subsection{Q1: Does \ours enable effective training of humanoids to perform diverse tasks in a human-like manner?}

We design the following tasks in simulation:
\begin{itemize}
\item[]  \textbf{Push Box}: Push a $30''\!\times\!40''\!\times\!40''$, 4\,kg box (friction 0.5–2.0).
\item[] \textbf{Kick Box}: Kick a $15''\!\times\!20''\!\times\!20''$, 0.5\,kg box.
\item[] \textbf{Lift Box}: Lift a $15''\!\times\!20''\!\times\!20''$, 0.5\,kg box.
\item[] \textbf{Reach Box}: Run to a $30''\!\times\!40''\!\times\!40''$ box.
\item[] \textbf{Large Kick}: Strongly kick a $15''\!\times\!20''\!\times\!20''$, 0.5\,kg box.
\item[] \textbf{Kick Ball}: Dribble a football.
\item[] \textbf{Balance Ball}: Balance with one foot on a football.
\item[] \textbf{Push Cube}: Push an $8''$ cube on a tabletop to target.
\end{itemize}

We evaluate each task using three metrics (TABLE~\ref{tab:sim}), with definitions provided in the caption. The choice of metrics follows the reward design of each task. For instance, in Push Box, the reward encourages forward motion along the x-axis while penalizing lateral drift; thus, we report Forward and Drift as metrics. Our framework achieves strong performance and successfully completes all tasks.
Our vision-based policy can push a 3.8-kilogram box, comparable in size to the robot, an average of 37 meters per minute. It can also dribble a football an average of 135 meters per minute, with 121 meters forward, indicating both straightness and robustness.
This performance stems from our hierarchical design, which decomposes the problem into training a low-level motion tracker and a high-level motion generator. 

\begin{figure}[tbp]
    \centering
    \includegraphics[width=1.0\linewidth]{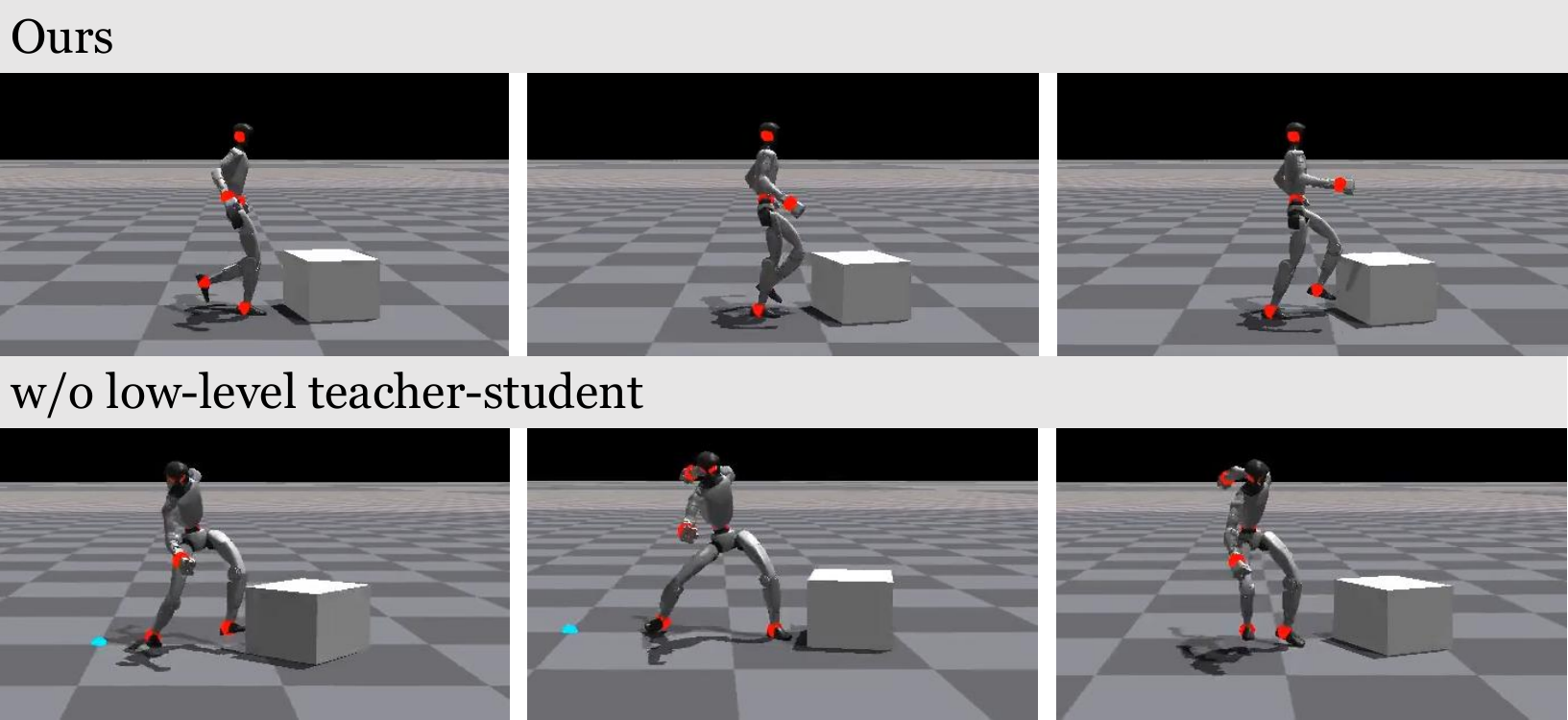}
    \captionsetup{type=figure}
    \caption{Box-kicking behaviors. With our teacher–student training (ours, top), the humanoid can mimic human-like motion, while training without it leads to non-human-like motion (bottom).} 
    \label{fig:kick_vis}
\end{figure}

\subsection{Q2: Can policies trained with \ours transfer robustly
to the real world?}

We deploy our visuomotor policies on the Unitree G1 humanoid robot equipped with its onboard RealSense D435i camera. The policies are evaluated on four tasks (Lift Box, Kick Ball, Kick Box, and Push Box) and all transfer successfully from simulation to the real world (Fig.~\ref{fig:real_world} and Fig. \ref{fig:push_box_real}). The robot can lift a box from the ground and maintain a stable hold, dribble a football forward with a human-like pose, kick a box in a coordinated manner, and push a large box straight and steadily. Beyond controlled laboratory settings, we also conducted outdoor experiments, in which the policies exhibit robustness to diverse real-world factors (Fig. \ref{fig:push_box_real}), including fluctuating lighting, irregular ground surfaces, and environmental distractions in the surroundings. These results confirm that our framework enables reliable deployment on physical hardware, allowing the humanoid to perform diverse whole-body loco-manipulation tasks. Supplementary videos are provided.

\subsection{Q3: How effectively does our framework demonstrate whole-body dexterity?}

\begin{figure}[tbp]
    \centering
    \includegraphics[width=1.0\linewidth]{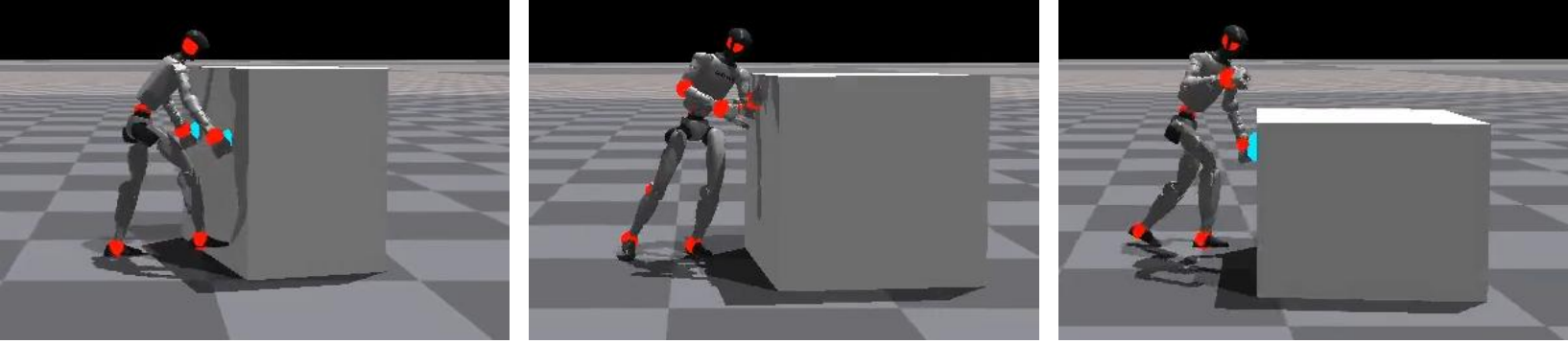}
    \captionsetup{type=figure}
    \vspace{-10pt}
    \caption{Whole-body dexterity in Push Box. Policies trained in different conditions exhibit distinct whole-body behaviors: on low-friction ground ($\mu=0.5$), it bends down to push with both hands; on high friction ($\mu=1.5$), it leans forward to push with its shoulder for greater force; when the box is lower, it switches to one-handed pushing while the other arm swings with its stride.}
    \label{fig:dex}
    \vspace{-10pt}
\end{figure}

Fig.~\ref{fig:real_world} and \ref{fig:sim} show that our method enables the robot to interact with objects using its entire body, including the lower limbs. We also observe flexible use of different body parts within the same task (Fig.~\ref{fig:dex}), demonstrating that our method enables adaptive whole-body strategies tailored to environment conditions. 

To investigate the source of this dexterity, we train a variant of \ours that uses only upper-body keypoints as commands instead of our 6-point design. This variant fails to effectively engage the feet, resulting in poor performance on tasks such as Balance Ball: it struggles to place the foot on top of the ball and instead consistently kicks it away (see supplementary videos).

\begin{table*}[tbp]
\centering 
\caption{Ablation study of \ours on three tasks in simulation. Experiment settings and metrics follow Table~\ref{tab:sim}. 
\textbf{Blind} denotes policies trained without visual input; here, \textbf{Ours w/o vision} is the student policy distilled without vision.}
\resizebox{1.0\textwidth}{!}{
\begin{tabular}{l|ccc|ccc|ccc}
\toprule
\textbf{Method} & \multicolumn{3}{c}{\textbf{Push Box}} & 
                \multicolumn{3}{c}{\textbf{Kick Box}} & 
                \multicolumn{3}{c}{\textbf{Kick Ball}} \\
\cmidrule(lr){2-4}\cmidrule(lr){5-7}\cmidrule(lr){8-10}
& Distance [m] $\uparrow$ & Forward [m] $\uparrow$ & Drift [m] $\downarrow$
& Distance [m] $\uparrow$ & Forward [m] $\uparrow$ & Drift [m] $\downarrow$
& Distance [m] $\uparrow$ & Forward [m] $\uparrow$ & Drift [m] $\downarrow$ \\
\midrule
\textbf{State-based} & & & & & & & & & \\
\quad Ours              & \cellcolor{myblue}\textbf{152 ± 36} & \cellcolor{myblue}\textbf{151 ± 29} & 13 ± 4  & 
\cellcolor{myblue}\textbf{78 ± 3} & \cellcolor{myblue}\textbf{78 ± 3} & \cellcolor{myblue}\textbf{0 ± 0}  &
\cellcolor{myblue}\textbf{189 ± 3} & \cellcolor{myblue}\textbf{189 ± 3} & 4 ± 1  \\
\quad w/o noise            & 2 ± 1 & 2 ± 1 & \cellcolor{myblue}\textbf{0 ± 0}  & 
30 ± 24 & 30 ± 20 & 1 ± 0  & 
136 ± 8 & 136 ± 7 & 4 ± 0  \\
\quad w/o clip             & 68 ± 118 & 67 ± 94 & 11 ± 16  & 
3 ± 5 & 3 ± 4 & \cellcolor{myblue}\textbf{0 ± 0}  & 
12 ± 15 & 12 ± 12 & 1 ± 1  \\
\quad DoF as Interface     & 10 ± 9 & 8 ± 6 & 5 ± 3  & 
40 ± 34 & 40 ± 28 & \cellcolor{myblue}\textbf{0 ± 0}  & 
0 ± 0 & 0 ± 0 & \cellcolor{myblue}\textbf{0 ± 0}  \\
\quad Local-Frame Tracker  & 38 ± 27 & 30 ± 23 & 16 ± 15  & 
45 ± 7 & 45 ± 5 & 1 ± 0  & 
109 ± 23 & 109 ± 19 & 7 ± 1  \\
\midrule
\textbf{Vision-based} & & & & & & & & & \\
\quad Ours              & \cellcolor{myblue}\textbf{37 ± 28} & \cellcolor{myblue}\textbf{19 ± 15} & 21 ± 12  & 
\cellcolor{myblue}\textbf{55 ± 5} & \cellcolor{myblue}\textbf{30 ± 3} & 33 ± 3   & 
\cellcolor{myblue}\textbf{135 ± 6} & \cellcolor{myblue}\textbf{121 ± 8} & 47 ± 12  \\
\quad w/o noise            & 2 ± 1 & 2 ± 1 & \cellcolor{myblue}\textbf{0 ± 0}  & 
25 ± 7 & 11 ± 4 & 15 ± 3  & 
86 ± 7 & 77 ± 7 & 30 ± 8   \\
\quad w/o clip             & 10 ± 18 & 9 ± 12 & 4 ± 5 & 
6 ± 7 & 5 ± 3 & 3 ± 3   & 
1 ± 1 & 0 ± 1 & \cellcolor{myblue}\textbf{0 ± 0}   \\
\quad  DoF as Interface    & 10 ± 2 & 6 ± 1 & 6 ± 1   & 
5 ± 4 & 1 ± 0 & 4 ± 3  & 
0 ± 0 & 0 ± 0 & \cellcolor{myblue}\textbf{0 ± 0}  \\
\quad Local-Frame Tracker  & 14 ± 11 & 7 ± 5 & 8 ± 6  & 
27 ± 15 & 16 ± 9 & 15 ± 6  & 
38 ± 17 & 34 ± 13 & 12 ± 4  \\
\quad Visual RL  & 25 ± 16 & 11 ± 6 & 16 ± 9  & 
0 ± 0 & 0 ± 0 & \cellcolor{myblue}\textbf{0 ± 0}  & 
0 ± 0 & 0 ± 0 & \cellcolor{myblue}\textbf{0 ± 0}  \\
\midrule
\textbf{Blind}  & & & & & & & & & \\
\quad  Ours w/o vision & 2 ± 0 & 2 ± 0 & 1 ± 0 & 
0 ± 0 & 0 ± 0 & 0 ± 0  & 
1 ± 0 & 1 ± 0 & 0 ± 0 \\
\bottomrule
\end{tabular}
}
\label{tab:ablation}
\end{table*}

\subsection{Q4: How well does \ours utilize vision for object interaction?}

All our tasks rely heavily on robot egocentric vision, where object positions are heavily randomized and robots can only perceive them via vision. To directly assess the role of vision, we train a variant of our policy distilled without visual input (TABLE~\ref{tab:sim}). This variant shows a substantial performance drop compared to its vision-enabled counterpart, highlighting the effectiveness of \ours in leveraging vision for robust and adaptive object interaction.

\subsection{Q5: Are design choices of \ours all necessary for the success of the system?}

\label{sec:ablation}

\begin{figure}[tbp]
    \centering

    \begin{subfigure}[t]{0.5\textwidth}
        \centering
        \includegraphics[width=\linewidth]{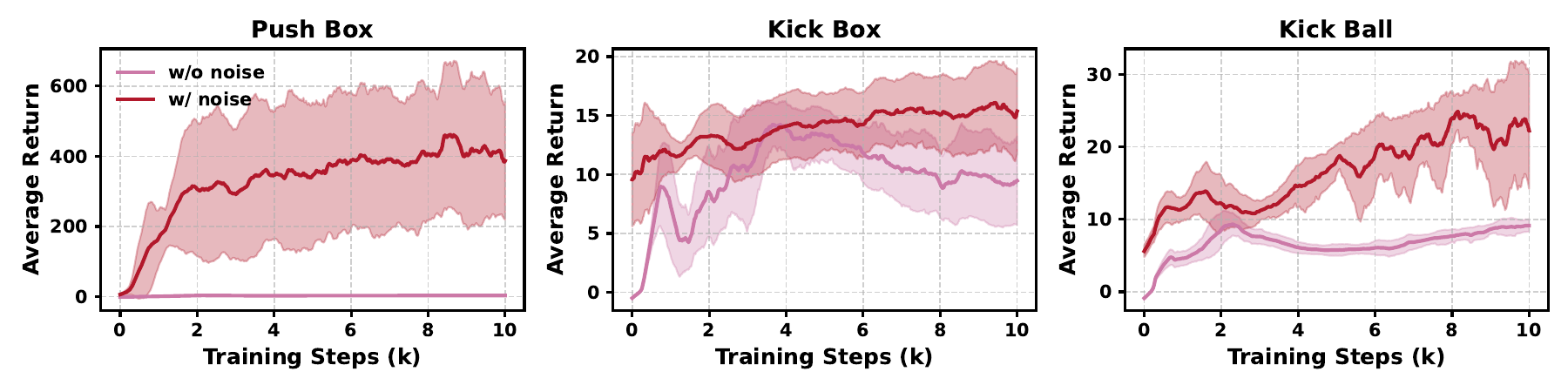}
        \vspace{-10pt}
        \caption{Ablation on the noise augmentation for keypoint tracker training.}
        \label{fig:ablate_noise}
    \end{subfigure}

    \begin{subfigure}[t]{0.5\textwidth}
        \centering
        \includegraphics[width=\linewidth]{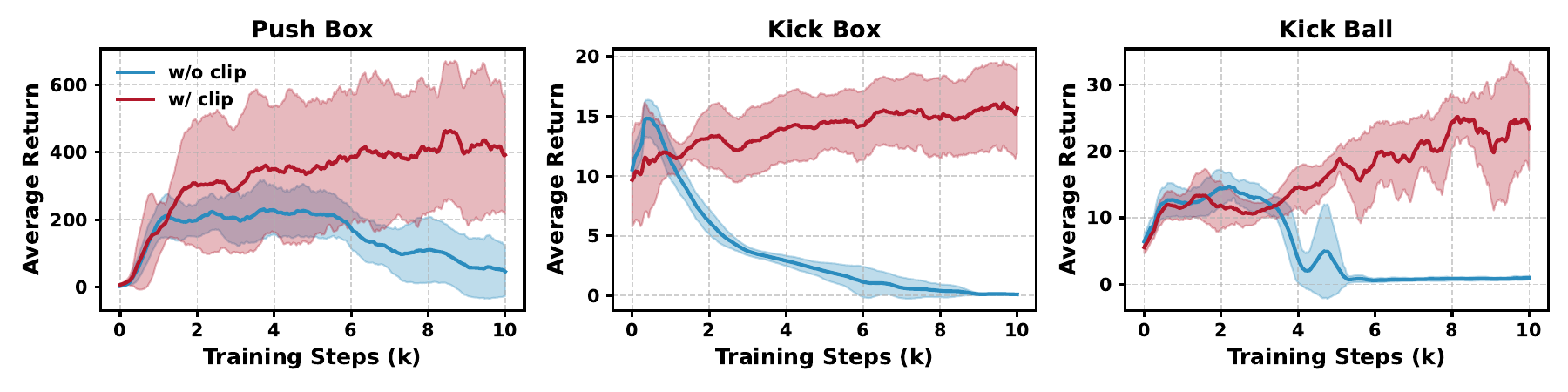}
        \vspace{-10pt}
        \caption{Ablation on action clip for the high-level policy.}
        \label{fig:ablate_clip}
    \end{subfigure}

    
    \begin{subfigure}[t]{0.5\textwidth}
        \centering
        \includegraphics[width=\linewidth]{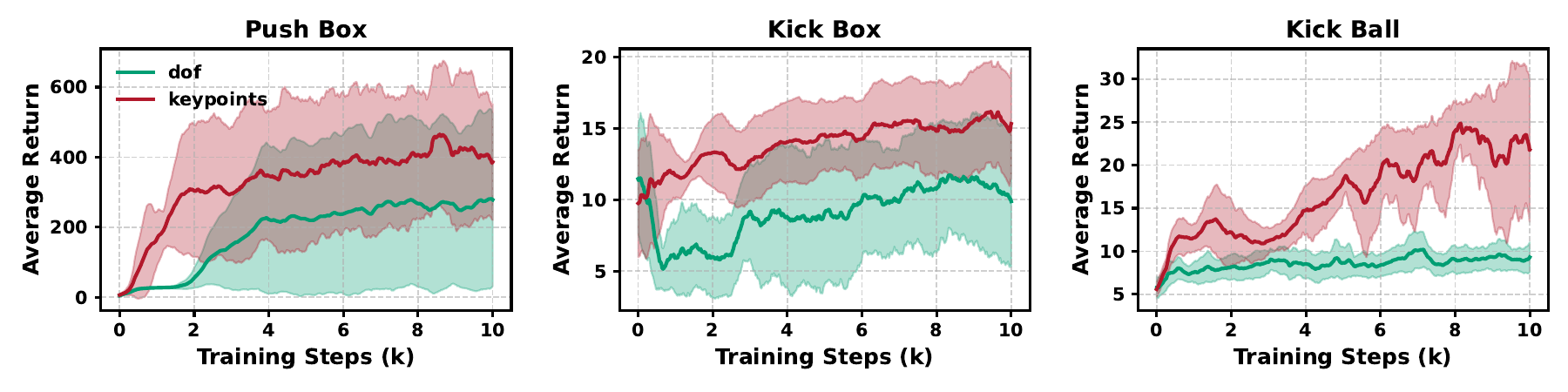}
        \vspace{-10pt}
        \caption{Ablation on the interface design.}
        \label{fig:ablate_dof}
    \end{subfigure}

    \caption{Training curves of our state-based keypoint generator compared with its variants.}
    \vspace{-10pt}
    \label{fig:ablate}
\end{figure}

To evaluate the contribution of key design components, we conduct extensive ablation studies on three tasks in simulation (TABLE~\ref{tab:ablation}). We compare against the following variants of \ours:
\begin{enumerate}
\item[] \textbf{w/o noise}: Distill a low-level keypoint tracker without adding noise to keypoint commands, then train the high-level policy on top.  
\item[] \textbf{w/o clip}: Use the same low-level keypoint tracker, but remove action clipping in the high-level policy.  
\item[] \textbf{DoF as Interface}: Distill a low-level tracker with a 23-DoF interface, then train the high-level policy on top.  
\item[] \textbf{Local-Frame Tracker}: Train the low-level motion and keypoint tracker with inputs and rewards computed in the local frame, then train the high-level policy on top.  
\item[] \textbf{Visual RL}: Directly train the visuomotor policy with RL for the same number of steps as the combined teacher–student training in the default setup.  
\end{enumerate}
For w/o noise and DoF as Interface, the tracker is distilled from the same teacher policy as in our method.
Moreover, for all baselines, only one component is altered while all others are kept identical. Removing multiple components (e.g., DoF interface without noise) leads to even worse performance than single-component variants. All methods share the same reward design and training parameters.

Metrics for the three tasks are reported in TABLE~\ref{tab:ablation}, with training curves in Fig.~\ref{fig:ablate}. It is observed that none of the baselines matches the performance of \ours.
We find that removing noise degrades performance in Kick Box and Kick Ball, and causes complete failure in Push Box (Fig.~\ref{fig:ablate_noise}). Eliminating action clipping leads to return collapse in later training stages across all tasks (Fig.~\ref{fig:ablate_clip}), likely because the high-level action space becomes excessively large. Using a DoF interface significantly reduces performance in Push Box and Kick Box, and prevents learning in Kick Ball (Fig.~\ref{fig:ablate_dof}). Training trackers in the local frame also results in substantial drops (TABLE ~\ref{tab:ablation}), as global-frame tracking better reduces drift and facilitates high-level training. Finally, Direct Visual RL fails entirely in Kick Box and Kick Ball.

  Besides, we find that removing teacher–student training for the low-level tracker, \textit{i.e.}, single-stage RL, produces highly unnatural behaviors during keypoint generator training (Fig.~\ref{fig:kick_vis}), underscoring the importance of the teacher-student framework for human-like keypoint tracking.
\section{Conclusions and Limitations}

In this work, we presented \ours, a visual sim-to-real framework that integrates visual perception with whole-body control for humanoid loco-manipulation and object interaction. By combining a high-level keypoint generator with a low-level general keypoint tracker, our approach enables humanoid robots to robustly perform diverse loco-manipulation tasks directly from visual and proprioceptive inputs. Experiments in both simulation and the real world demonstrate that \ours enables robust deployment from simulation to the real robot.

 \noindent\textbf{Limitations.} While our hierarchical design generalizes across a range of loco-manipulation tasks, more complex interactions involving deformable objects or human collaboration remain unexplored. Besides, though sim-to-real transfer has been effective in our tested scenarios, further scaling to long-horizon tasks and diverse real-world environments may require additional advances in domain randomization and adaptive control. We leave these directions for future research.

 \section*{Acknowledgments}
 We would like to thank all members of the CogAI group and The Movement Lab from Stanford University for their support. We also thank the Stanford Robotics Center for providing the experiment space. This work is in part supported by Stanford Institute for Human-Centered AI (HAI), Stanford Robotics Center (SRC), ONR MURI N00014-22-1-2740, ONR MURI N00014-24-1-2748, and  NSF:FRR 215385.
 
\bibliographystyle{IEEEtran}
\bibliography{main}

\end{document}